\documentclass{article}
\PassOptionsToPackage{numbers, compress}{natbib}
\usepackage[preprint]{neurips_2026}

\usepackage[utf8]{inputenc}
\usepackage[T1]{fontenc}
\usepackage{amsfonts}
\usepackage{amssymb}
\usepackage{amsmath}
\usepackage{mathtools}
\usepackage{array}
\usepackage{booktabs}
\usepackage{graphicx}
\usepackage{microtype}
\usepackage{multirow}
\usepackage{nicefrac}
\usepackage{url}
\usepackage[dvipsnames,x11names]{xcolor}
\usepackage{xspace}
\usepackage{fontawesome5}

\newcommand{\method}{\textup{\textsc{Instruct-Particulate}}\xspace}
\newcommand{\ie}{i.e.\xspace}
\newcommand{\eg}{e.g.\xspace}
\newcommand{\vs}{vs.\xspace}
\newcommand{\best}[1]{\textcolor{Maroon}{\textbf{#1}}}
\newcommand{\secondbest}[1]{\textcolor{RoyalBlue}{\textbf{#1}}}

\definecolor{Urlcolor}{RGB}{251,111,146}
\definecolor{Linkcolor}{RGB}{193,18,31}
\definecolor{CiteColor}{RGB}{32,126,190}
\definecolor{darkgreen}{RGB}{0,128,0}
\usepackage[
  pagebackref,
  breaklinks,
  colorlinks,
  urlcolor=Urlcolor,
  linkcolor=Linkcolor,
  citecolor=CiteColor
]{hyperref}
\usepackage[capitalize]{cleveref}

\makeatletter
\renewcommand{\paragraph}{%
  \@startsection{paragraph}{4}%
  {\z@}{-0.5em}{-0.5em}%
  {\normalfont\normalsize\bfseries}%
}
\makeatother

\title{\method{}: Scaling Feed-Forward 3D Object Articulation with Kinematic Control}

\author{%
  \textbf{Ruining Li}$^{1*}$ \quad
  \textbf{Yuxin Yao}$^{2*}$ \quad
  \textbf{Matt Zhou}$^{2}$ \quad
  \textbf{Chuanxia Zheng}$^{3}$ \quad \\[0.2em]
  \textbf{Christian Rupprecht}$^{1}$ \quad
  \textbf{Joan Lasenby}$^{2}$ \quad
  \textbf{Shangzhe Wu}$^{2\dag}$ \quad
  \textbf{Andrea Vedaldi}$^{1\dag}$ \\[0.5em]
  $^{1}$University of Oxford \quad
  $^{2}$University of Cambridge \quad
  $^{3}$Nanyang Technological University \\[0.5em]
  \href{http://instruct-particulate.github.io/}{\texttt{instruct-particulate.github.io}}
}

\begin{document}
\maketitle
\begingroup
\renewcommand{\thefootnote}{}
\footnotetext{$^*$Equal contribution. $^\dag$Equal advising.}
\endgroup

\begin{figure}[h]
\vspace{-0.1in}
\centering
\includegraphics[trim={0 0bp 0bp 0}, clip, width=\linewidth]{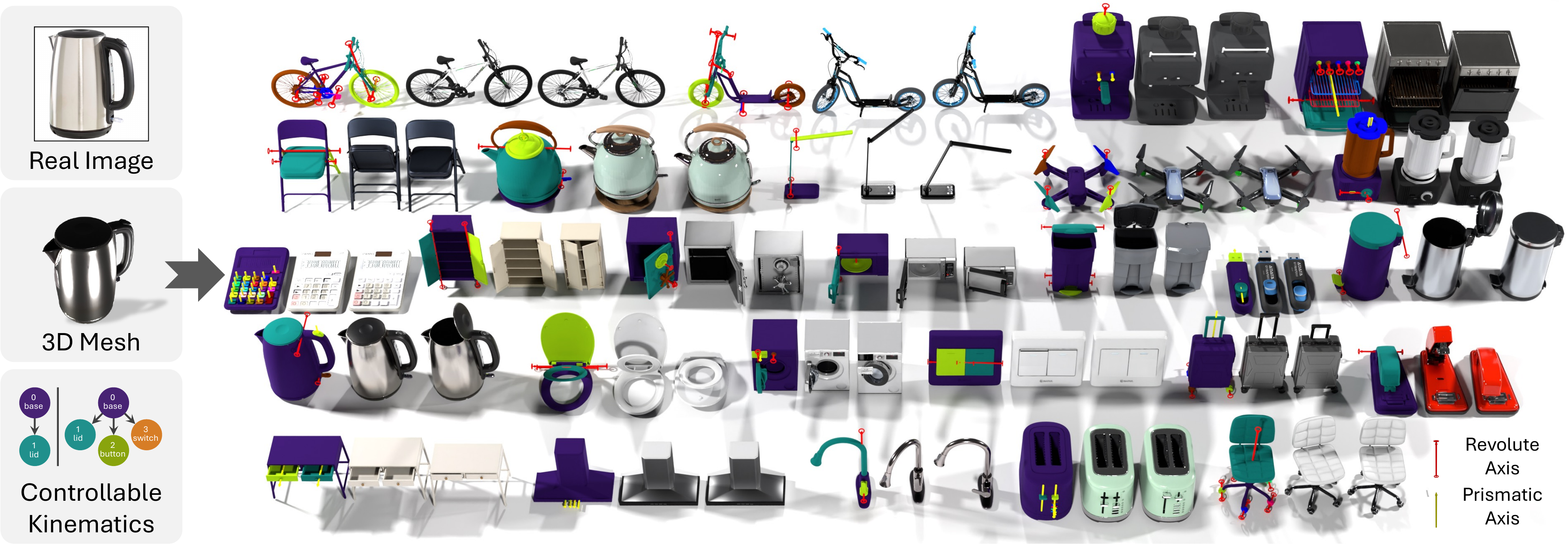}
\vspace{-1.6em}
\caption{
\textbf{Articulated 3D objects predicted by \method from real-world images}.
Our model infers articulated structures from static 3D assets, including outputs from off-the-shelf 3D generators, and supports optional kinematic prompting.
This allows for generating diverse, realistic articulated 3D objects directly from real-world images.
}%
\label{fig:teaser}
\end{figure}

\begin{abstract}
Reconstructing articulated 3D objects is important for animation, gaming, and robotic simulations.
Recent neural networks can estimate the articulated structure of 3D objects, but their generalization remains limited by the scarcity of annotated data for this task.
To address this gap, we introduce \emph{\method}, a model that takes a 3D mesh together with a target kinematic specification, including part descriptions, connectivity, joint types, and optional point prompts, and predicts the corresponding kinematic part segmentation and joint motion parameters.
The kinematic specification disambiguates the task and allows the model to target annotations of different granularity, thereby making it possible to use more abundant heterogeneous training data.
At test time, the kinematic specification can be obtained automatically from large-scale vision-language models, so the model can be applied to any input mesh.
To train our model at scale, we construct a heterogeneous dataset of more than $150{,}000$ articulated 3D objects, extending existing publicly available collections with data obtained by partially labelling other 3D models (monolithic or already decomposed into parts) with kinematic labels by means of vision-language models.
Experiments show that our model generalizes better across categories and to AI-generated meshes, enabling articulated asset reconstruction from real-world images via image-to-3D models.
\end{abstract}

\section{Introduction}%
\label{sec:intro}

Understanding and manipulating articulated objects is often necessary to interact with the physical world, and thus a key capability that physical agents must acquire.
To this end, we consider the problem of reconstructing the articulated structure of a 3D object, decomposing it into parts with their articulation parameters (\cref{fig:teaser}).
Prior work on this problem~\cite{li2026particulate,li2026art,cao2025physx3d,li2026monoart,wang2026artllm} has so far relied only on small training datasets~\cite{xiang20sapien:}, thus missing the benefits of large-scale pre-training already demonstrated in text~\cite{brown20language}, image~\cite{deng09imagenet:,schuhmann2022laion,rombach22high-resolution, radford21learning}, video~\cite{brooks24video,seedance2026seedance}, and 3D~\cite{deitke23objaverse:,objaverseXL} understanding.

The lack of high-quality articulated 3D data remains a major bottleneck for further progress in this area.
In particular, while recent works have increased the number of available articulated 3D models via procedural generation~\cite{joshi25infinigen-sim:,lian25infinite,cao2025physx3d}, part permutation~\cite{cao2025physx3d,li2026art}, and manual annotation~\cite{gao25meshart:,physxanything}, the diversity of the resulting data is lacking.
As a consequence, models trained on such data struggle to generalize to novel objects and categories, as we demonstrate in \cref{sec:exp}.

In this work, we introduce \emph{\method}, an approach that significantly boosts generalization in articulated object understanding.
To achieve this, we first focus on scaling both data \emph{size} and \emph{diversity}.
We are motivated by the observations that
(1) there are several orders of magnitude more generic 3D assets~\cite{deitke23objaverse:,objaverseXL} than assets paired with articulation annotations, and
(2) frontier vision-language models (VLMs)~\cite{openai26gpt54,googledeepmind2026gemini,bai2025qwen3vl} possess strong 2D understanding of object articulation.
We thus propose a category-agnostic data engine that uses the reasoning and generation capabilities of an off-the-shelf VLM~\cite{raisinghani2025nanobananapro} to pseudo-label (real or generated) 3D assets with articulated parts.
Using this engine, we extract articulated part segmentations for $27$k synthetic 3D assets of everyday objects spanning $432$ categories.
We further incorporate $120$k curated 3D assets with generic part decompositions and $10$k articulated 3D assets across $200$ categories produced by a coding agent specialized for articulated 3D design~\cite{zhou2026articraft}.

This expanded data mixture improves the potential for generalization beyond prior work, but it also introduces a new challenge:
often there is no single correct way to assign an articulated structure to a given 3D object, and greater data diversity therefore introduces inconsistencies in part granularity and semantics.
A naive model would ``average'' over multiple plausible annotations and produce suboptimal results.

We address this problem by proposing a new formulation in which the model is \emph{instructed} to extract a particular articulated structure, specified by an explicit kinematic structure (\ie, a list of parts and their connectivity), the joint types (\eg, revolute or prismatic), and optional 3D point prompts.
All of these pieces of information, which can be provided manually or extracted automatically by a VLM at test time, disambiguate the target articulated structure and enable the model to predict crisp, coherent 3D part segmentations.

This new model is implemented using a scalable encoder-decoder architecture.
Given a point cloud approximating the mesh together with kinematic prompts, the model predicts part labels for arbitrary surface query points and estimates the corresponding joint motion parameters.
At test time, it infers the articulated structure of new objects efficiently, in a feed-forward manner.
Furthermore, while we focus on the \emph{analysis} of given 3D objects, we also consider \emph{synthesizing} them from scratch by leveraging off-the-shelf 3D generators~\cite{lai2025hunyuan3d25highfidelity3d,xiang25native} in combination with our model.

We summarize our contributions as follows:
(1) We propose \method, a state-of-the-art model that takes as input a static 3D object together with kinematic instructions, and outputs a corresponding fully articulated 3D object in a feed-forward manner;
(2) We design a pipeline to pseudo-label a large library of 3D models with 3D articulated parts;
(3) We show that our model enables the generation of diverse articulated 3D objects from real-world images, producing assets that are directly exportable to physics simulators.

\section{Related Work}%
\label{sec:related}

\paragraph{Reconstruction and generation of articulated 3D objects.}

Early approaches for reconstructing 3D articulated objects assume dense multi-view inputs and use test-time optimization~\cite{liu23paris:,song24reacto:,wei22self-supervised,mu2021sdf,chen25freeart3d:,liu25building,liu25videoartgs:}, which makes them scale poorly.
More recent approaches train feed-forward models that generate articulated 3D assets from few images~\cite{lei2023nap,chen24urdformer:,liusingapo,cao2025physx3d,li2025urdfanything,physxanything,wu2025dipo,wang2025kinematify,li2026art,liu2026pact,wu2026urdfanythingplus,li2026monoart,mandi2025real2code},
but their effectiveness is often constrained by limited training data.
To improve generalization, several researchers have proposed to leverage foundation vision models~\cite{li24dragapart,li25puppet-master,lu25dreamart:} to infer plausible part decompositions and motion constraints.
However, these methods often struggle with small parts and cannot predict multiple parts in parallel.
In this work, inspired by~\cite{sutton19the-bitter}, we scale the size and diversity of the training data, obtaining a model which is much more robust and general.

\paragraph{Feed-forward 3D part segmentation and articulation estimation.}

We are motivated by recent progress in 3D part segmentation, where state-of-the-art methods have moved beyond lifting masks from 2D foundation models such as SAM~\cite{kirillov23segment} and GLIP~\cite{li22grounded} in~\cite{liu23partslip:,zhou23partslip:,xue25zerops:,abdelreheem23satr:,tang24segment}
toward data-driven models that operate directly in 3D~\cite{chen25partgen,chen25autopartgen,liu2025partfield,ma2025p3sam,ma25find,yang2024sampart3d}.
This shift has been enabled by increasingly large and diverse 3D datasets with part-level annotations.
Several works extend these models to jointly predict articulation and rigging~\cite{wu23magicpony,jakab24farm3d,li24learning,liu2025riganything,deng2025anymate,song2025magicarticulate,li2026particulate}.
However, applicable datasets are dominated by humanoid and animal assets commonly used in games.
As a result, these methods transfer poorly to other common objects, which are the focus of this work.

\paragraph{3D datasets.}

Existing 3D object datasets have been obtained by scanning real objects~\cite{downs22google,collins22abo:,wu23omniobject3d:} or downloading manually-authored 3D assets from the web~\cite{chang15shapenet,deitke23objaverse:,objaverseXL}.
While some of these collections are large, they generally lack information about articulation.
Some works exploit `accidental metadata', such as the fact that manually-authored 3D assets are already organized into different components, to derive part annotations, but such decompositions rarely align with kinematic parts.
Conversely, articulated 3D datasets~\cite{xiang20sapien:,liu22akb-48:,geng23gapartnet:,wang2024grutopia,cao2025physx3d,nasiriany26robocasa365:} are typically annotated by hand, which limits their scale and diversity.
Procedural generation has also been used to expand articulated 3D asset collections~\cite{joshi25infinigen-sim:,lian25infinite,cao2025physx3d}.
However, these rule-based pipelines are difficult to scale to the long tail of real-world objects, leaving them practical only for a limited set of categories.
In this work, we introduce complementary strategies for scaling this data.

\section{Building a Large Dataset of Articulated 3D Objects}%
\label{sec:scaling-data}

\begin{figure}[t]
\centering
\includegraphics[width=\linewidth]{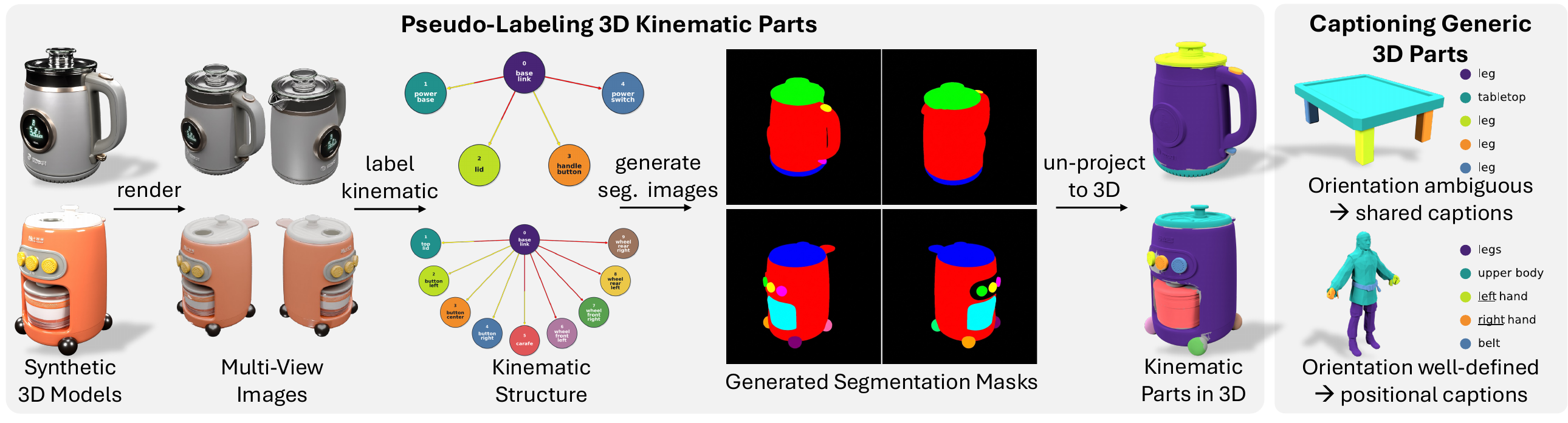}
\vspace{-1.6em}
\caption{
\textbf{Data pipelines of \method}.
\emph{Left}: synthetic 3D assets are first rendered from multiple views.
A vision-language model then extracts their kinematic structures and generates segmentation masks according to a fixed color scheme.
The 2D segmentation results are then unprojected to 3D.
Certain occluded regions are left unlabeled (visualized in gray).
\emph{Right}: for assets with existing part decompositions, we generate part captions that avoid arbitrary spatial labels when the object orientation is ambiguous, but use positional cues when a canonical orientation is available.
}%
\label{fig:pseudo-labeling}
\end{figure}

A key contribution of our paper is to build a large dataset of articulated 3D objects for training.
To this end, we experiment with three complementary approaches.
First, we build a VLM-based pipeline to partially label a large library of static 3D objects (\cref{sec:pseudo-labeling}).
Second, for 3D objects that are already decomposed into parts, we adapt the VLM-based pipeline to assign textual captions to each part (\cref{sec:generic-part-segmented}).
Third, we leverage a recent 3D coding agent to generate additional articulated objects (\cref{sec:articulated-3d-models-with-coding-agents}) with full supervision of their joint parameters.

\paragraph{Articulated 3D objects.}

We first specify what we mean by an articulated 3D object.
An object is given by a 3D surface $M \subset \mathbb{R}^3$ (a mesh in practice).
The mesh has $P$ \emph{kinematic parts} defined by a mapping $m : M \mapsto [P]=\{1,\dots,P\}$ that assigns each point $\mathbf{x} \in M$ of the surface to a part $m(\mathbf{x}) \in [P]$.
Parts compose in a \emph{kinematic structure}
$
\mathcal{K} \coloneqq \{(u_e,v_e,\tau_e)\}_{e=1}^{P-1},
$
which is a collection of $P-1$ joints where part $u_e\in [P]$ connects to part $v_e \in [P]$ with a joint of type
$
\tau_e \subset \{ \text{pri}, \text{rev}\}
$
(\ie, fixed, revolute, prismatic, or both).
The kinematic structure forms a directed tree, and only specifies the connectivity and motion types of the parts.
We further associate to each movable joint $e$ (\ie, $\tau_e \neq \varnothing$) an \emph{axis}
$
\mathbf{a}_e = (
    \mathbf{d}_e,\mathbf{p}_e
) \in
\mathbb{S}^2 \times \mathbb{R}^3
$
comprising a unit direction $\mathbf{d}_e$ and a pivot point $\mathbf{p}_e$ with motion bounds
$
[\theta_e^-,\theta_e^+] \subset \mathbb{R}
$.
A joint that is both revolute and prismatic can have separate axes for rotation and translation.
An articulated 3D object is then defined as a tuple
$
(M,m,\mathcal{K},\mathbf{a},\theta)
$
comprising a mesh, part assignments, kinematic structure, and joint parameters.

To specify instructions to our model, we further define the \emph{kinematic condition} as 
$
C 
\coloneqq \left(
\{(t_p,\mathbf{x}_p)\}_{p=1}^{P}, 
\mathcal{K}\right),
$
where $\mathcal{K}$ is the kinematic structure defined above, and each part $p$ is associated with a text prompt $t_p$ and an optional point prompt $\mathbf{x}_p \in M$ such that $m(\mathbf{x}_p)=p$ (\ie, a 3D point that belongs to the part).

\subsection{Pseudo-Labeling 3D Articulated Parts with Vision-Language Models}%
\label{sec:pseudo-labeling}

Our first approach to obtain articulated objects is to start with a large library of synthetic 3D models $M$ and use an off-the-shelf vision-language model (VLM) to pseudo-label them with articulated part segmentations, obtaining a corresponding part segmentation map $m$, kinematic structure $\mathcal{K}$, part captions $\{t_p\}_{p=1}^P$, and point prompts $\{\mathbf{x}_p\}_{p=1}^P$ (which can be easily obtained by sampling $m$).

Given a 3D object $M$ (\ie, a textured mesh), the first step is to extract 3D parts from it.
While there are numerous 3D part segmentation methods~\cite{chen25autopartgen,chen25partgen,yang2024sampart3d,liu2025partfield,ma2025p3sam} and datasets~\cite{Mo_2019_CVPR,wang2025partnext,dong2025contextual,hunyuan3d2026hy3dbench}, these focus on \emph{semantic} part decompositions, which often do not correspond well to object articulation.

Instead, we use a VLM~\cite{googledeepmind2026gemini} to extract the names of articulated parts and their connectivity from multi-view renderings of each 3D object $M$, thus defining $\mathcal{K}$.
We then prompt an image generator~\cite{raisinghani2025nanobananapro} to assign a different color to each kinematic part, thus obtaining an instance segmentation map for each view.
Since the available parts are known a priori from the kinematic labeling, we instruct the model to use a fixed color scheme of our choice, which makes it easy to identify the parts in all views.
We then obtain 2D segmentations by assigning each pixel to the part with the closest color according to the prompt.
Finally, we project the 2D segmentations back to the 3D model to obtain the 3D articulated part segmentations $m$.
The entire pipeline is illustrated in \cref{fig:pseudo-labeling} (left).

We label AI-generated assets from HY3D-Bench-Synthetic~\cite{hunyuan3d2026hy3dbench} using this pipeline, excluding objects that are either fully rigid or exhibit soft, non-rigid deformations.
This yields $27$k synthetic 3D assets spanning $432$ categories.

\subsection{Augmenting Part-Segmented 3D Models}%
\label{sec:generic-part-segmented}

In addition to starting from static 3D models $M$, we also consider datasets like HY3D-Bench-Part-Level~\cite{hunyuan3d2026hy3dbench} that contain part-segmented objects $(M,m)$.
While the parts annotated in these datasets are not necessarily kinematic, we posit that they still provide useful supervision for aligning the part text prompts $\{t_p\}_{p=1}^P$ with the geometry $M$.

We obtain these prompts $t_p$ by captioning each of the $P$ parts using a VLM, following a pipeline similar to \cref{sec:pseudo-labeling}.
The VLM is provided with a rendering of the textured 3D model and a segmentation map rendered from the same viewpoint, and outputs a set of candidate captions for each visible part.
A key design choice is how to deal with semantically identical parts that are hard to distinguish with text alone (e.g., the four legs of a dining table).
Inconsistent positional labeling (\eg, left \vs right, front \vs back) would likely confuse the model.
As shown in \cref{fig:pseudo-labeling} (right), we address this by carefully prompting the VLM to
(1) use spatial cues only when the object has a meaningful canonical orientation (\eg, the humanoid character); or otherwise
(2) output the exact same set of captions for all semantically identical parts (\eg, the table).
For parts with identical captions, we therefore keep their point prompts $\mathbf{x}_p$ (\ie, no random dropout) during training, enabling the model to distinguish them geometrically.

When selecting assets from HY3D-Bench-Part-Level~\cite{hunyuan3d2026hy3dbench}, we filter out assets with more than $10$ visible parts, as dense segmentation maps make it difficult for the VLM to reliably distinguish colors and assign accurate captions.
We caption the remaining $117$k objects and include them in training.

\subsection{Generating Articulated 3D Models with Coding Agents}%
\label{sec:articulated-3d-models-with-coding-agents}

The datasets above are labeled with part segments and part prompts $(M, m, \{t_p\}_{p=1}^P)$, but they do \emph{not} contain the joint parameters $J=(\mathbf{a}, \theta)$.
To supply our model with this supervision, we further consider $10$k articulated 3D objects spanning $200$ categories generated by Articraft~\cite{zhou2026articraft}, a recently introduced 3D coding agent.

\section{Model Architecture}%
\label{sec:model}

Having introduced our data, we now present the architecture of our model \method.
The goal of the model is to predict the articulated structure of a given 3D object following the directives given as an additional kinematic prompt.
Specifically, the input is a mesh $M$ together with a kinematic condition $C$ that specifies the target kinematic tree $\mathcal{K}$ and the part prompts $\{(t_p,x_p)\}_{p=1}^P$.
The output is a map $m$ that assigns each point on the surface of $M$ to a part, as well as the motion parameters $J=(\mathbf{a},\theta)$ of the movable joints in $\mathcal{K}$.

\begin{figure}[t]
\centering
\includegraphics[width=\linewidth]{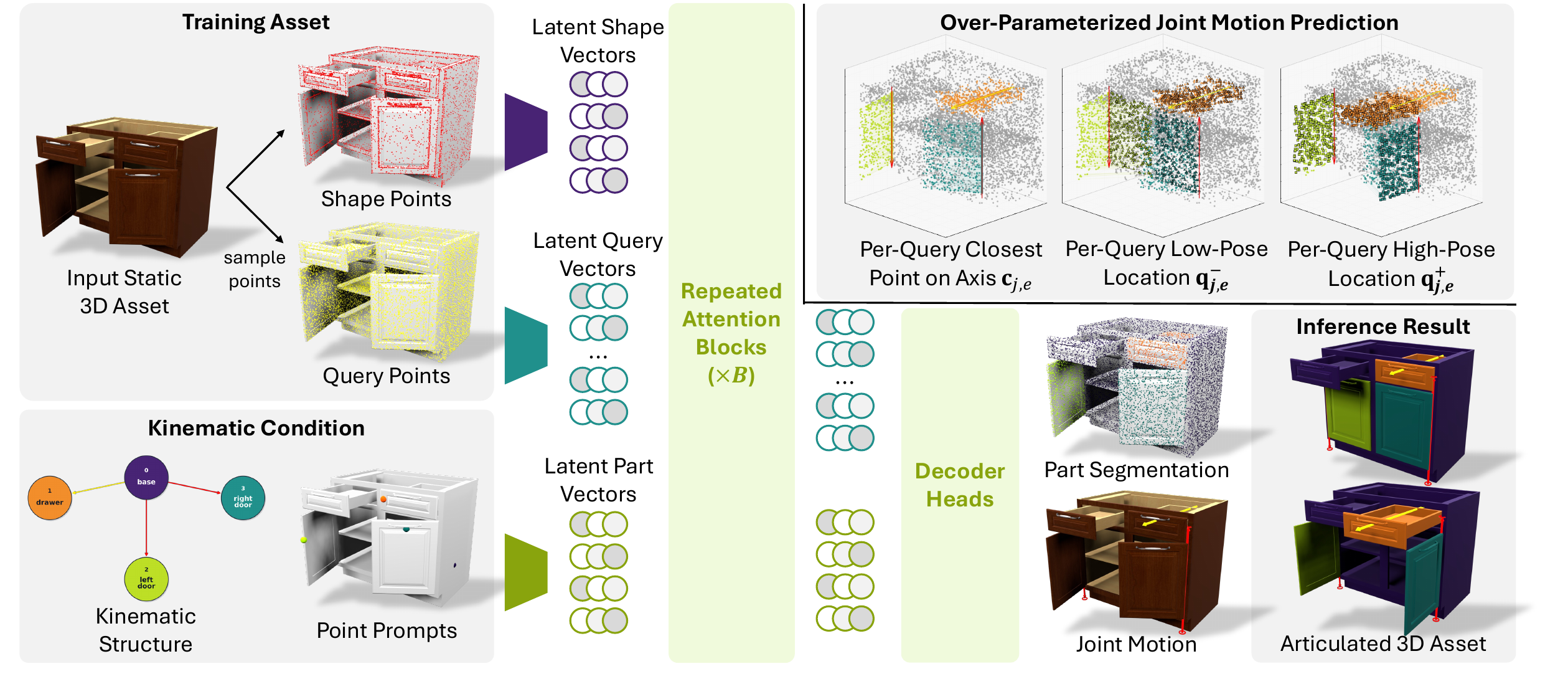}
\vspace{-1.6em}
\caption{
\textbf{Architectural overview of \method}.
Given a 3D mesh represented as a surface \textcolor{red}{point cloud} and a description of the desired kinematic structure, the model predicts part associations for arbitrary surface \textcolor{Yellow4}{query points} and motion parameters for the joints.
Internally, shape, part, and query tokens are processed by $B$ attention blocks before lightweight decoder heads recover segmentation and joint parameters.
\emph{Top-Right}: Joint motion is decoded from over-parameterized predictions of each query point's closest point on the joint axis and its locations at the joint limits.
}%
\label{fig:architecture}
\end{figure}

Next, we describe the architecture, illustrated in \cref{fig:architecture}.
We begin by explaining how the model encodes the various inputs as tokens (\cref{sec:architecture-encoder}), followed by the attention blocks used to process these tokens (\cref{sec:architecture-attention}), and finally the decoder heads that predict the part segmentation and joint motion parameters (\cref{sec:architecture-decoder}).

\subsection{Encoders}%
\label{sec:architecture-encoder}

\paragraph{Shape Tokens.}

The model takes as input a shape $M$ which, inspired by~\cite{liu2025partfield,ma2025p3sam,zhang20233dshape2vecset,li2026particulate}, is represented by sampling $N$ points
$
S \coloneqq \{\mathbf{x}_i \in \mathbb{R}^3\}_{i=1}^N \subset M
$
on it.
As in~\cite{li2026particulate}, we encode each point with its associated surface normal $\mathbf{n}_i \in \mathbb{R}^3$ and feature vector $\mathbf{f}_i \in \mathbb{R}^d$ obtained using PartField~\cite{liu2025partfield}, and sum them to obtain the point token
$
    \tilde{\mathbf{x}}_i
    =
    \operatorname{embed}(\mathbf{x}_i,\mathbf{n}_i,\mathbf{f}_i)
    =
    \phi_x(\mathbf{x}_i)
    +
    \phi_n(\mathbf{n}_i)
    +
    \phi_f(\mathbf{f}_i)
    \in \mathbb{R}^D
$.
Here, $\phi_x$, $\phi_n$, and $\phi_f$ are separate MLP embedders.
Following VecSet~\cite{zhang20233dshape2vecset}, these $N$ point tokens are then reduced to a much smaller number $L \ll N$ of \emph{shape tokens} $\mathbf{Z}^{S} \in \mathbb{R}^{L \times D}$.
This is achieved by letting a fixed set of learnable embeddings $\mathbf{Z}^{0} \in \mathbb{R}^{L \times D}$ cross attend the point tokens $\{\tilde{\mathbf{x}}_i\}_{i=1}^{N}$, i.e.,
$
    \mathbf{Z}^{S}
    =
    \operatorname{CrossAttn}
    \left(
        \mathbf{Z}^{0},
        \{\tilde{\mathbf{x}}_i\}_{i=1}^{N}
    \right)
$.

\paragraph{Point Query Tokens.}

In order to express the part segmentation, as well as some of the articulation parameters, we further consider additional \emph{query points}
$
Q \coloneqq \{\mathbf{q}_j \in \mathbb{R}^3\}_{j=1}^K \subset M
$
sampled on the shape and embed them in the same way as before, namely
$
\tilde{\mathbf{q}}_j^{0}
=
\operatorname{embed}(\mathbf{q}_j,\mathbf{n}_j,\mathbf{f}_j)
$.

\paragraph{Part Tokens.}

The model is also given a sequence of part descriptors $\{(t_p,\mathbf{x}_p)\}_{p=1}^{P}$, where $t_p$ is a text prompt describing part $p$ and $\mathbf{x}_p$ is an optional point that belongs to that part.
We associate a token $\tilde{\mathbf{l}}_p^{0} \in \mathbb{R}^D$ to each part $p$ by encoding the text prompt $t_p$ with an MLP applied to its CLIP~\cite{radford21learning} embedding, and again reusing the point embedder $\operatorname{embed}(\cdot)$ for the point prompt $\mathbf{x}_p$, so that
$
    \tilde{\mathbf{l}}_p^{0}
    =
    \phi_t(\operatorname{CLIP}(t_p))
    +
    \operatorname{embed}(\mathbf{x}_p,\mathbf{n}_p,\mathbf{f}_p)
    \in \mathbb{R}^D
$.

\subsection{Attention}%
\label{sec:architecture-attention}

The shape, query, and part tokens are then processed by a transformer~\cite{vaswani17attention} using a custom attention pattern.
The part tokens are meant to infer the articulation parameters of each part with respect to its parent.
To this end, each of $B$ transformer blocks updates the part tokens $\tilde{\mathbf{l}}_p$ via self-attention and by cross-attending the shape tokens $\mathbf{Z}^{S}$ to obtain the necessary information on the shape of the object:
$$
\{\tilde{\mathbf{l}}_p^{b}\}_{p=1}^{P}
=
\operatorname{SelfAttn}
\left(
\operatorname{CrossAttn}
(
    \{\tilde{\mathbf{l}}_p^{b-1}\}_{p=1}^{P}, \mathbf{Z}^{S}
)
\right),
\quad
b = 1,\dots,B.
$$
The query tokens, on the other hand, are meant to infer the association of each query surface point to the corresponding part.
They are updated by cross-attending the shape tokens $\mathbf{Z}^{S}$ to obtain information on the object shape, and to the (updated) part tokens $\tilde{\mathbf{l}}_p$ to obtain information on the parts:
$$
\{\tilde{\mathbf{q}}_j^{b}\}_{j=1}^{M}
=
\operatorname{CrossAttn}
\left(
\operatorname{CrossAttn}(
    \{\tilde{\mathbf{q}}_j^{b-1}\}_{j=1}^{M}, \mathbf{Z}^{S}
),
\{ \tilde{\mathbf{l}}_p^{b} \}_{p=1}^{P}
\right),
\quad
b = 1,\dots,B.
$$
Crucially, we do not use self-attention between query tokens, for speed, and to avoid making the result dependent on the number of query points decoded together~\cite{li2025vanishing}.

\subsection{Decoders}%
\label{sec:architecture-decoder}

Once the final versions of the part $\tilde{\mathbf{l}}_p^{B}$ and query $\tilde{\mathbf{q}}_j^{B}$ tokens are obtained, decoder heads extract the part segmentation and joint motion parameters.

\paragraph{Part segmentation.}

For part segmentation, an MLP $\psi_{\mathrm{part}}$ scores each query-part pair, producing logits
$
\tilde{\mathbf{S}} \in \mathbb{R}^{M \times P}
$
with
$
\tilde{\mathbf{S}}_{j,p} = \psi_{\mathrm{part}}(\tilde{\mathbf{q}}_j^{B}, \tilde{\mathbf{l}}_p^{B})
$.

\paragraph{Motion parameters.}

For each movable joint $(u_e,v_e,\tau_e \neq \varnothing)$ in $\mathcal{K}$, we predict its axis of motion $\mathbf{a}_e = (\mathbf{d}_e,\mathbf{p}_e) \in \mathbb{S}^2 \times \mathbb{R}^3$, comprising a unit direction $\mathbf{d}_e$ and a pivot point $\mathbf{p}_e$,
along with lower and upper bounds $[\theta_e^-,\theta_e^+] \subset \mathbb{R}$.
Following~\cite{li2026particulate}, we predict these motion parameters in a per-query-point, over-parameterized manner.
Concretely, for each query point $\mathbf{q}_j$ of the child part $v_e$ and each allowed motion type $\tau \in \tau_e$,
an MLP $\psi_\text{joint}$ predicts an over-parameterized motion target
\[
    \left(
        \tilde{\mathbf{d}}_{j,e}^{\tau},
        \tilde{\mathbf{c}}_{j,e}^{\tau},
        \tilde{\mathbf{q}}_{j,e}^{\tau,-},
        \tilde{\mathbf{q}}_{j,e}^{\tau,+}
    \right)
    =
    \psi_\text{joint}
    \left(
        \tilde{\mathbf{q}}_j^{B},
        \tilde{\mathbf{l}}_{u_e}^{B},
        \tilde{\mathbf{l}}_{v_e}^{B},
        \mathbf{e}_{\tau}
    \right) \in \mathbb{S}^2 \times \mathbb{R}^3 \times \mathbb{R}^3 \times \mathbb{R}^3,
\]
where $\mathbf{e}_{\tau}$ is a learnable embedding that encodes the motion type $\tau$.
The output tuple of the MLP consists of a local axis-direction estimate $\tilde{\mathbf{d}}_{j,e}^{\tau}$, the closest point $\tilde{\mathbf{c}}_{j,e}^{\tau}$ on the joint axis to $\mathbf{q}_j$,
and the query point's limit-pose locations $\tilde{\mathbf{q}}_{j,e}^{\tau,-}$ and $\tilde{\mathbf{q}}_{j,e}^{\tau,+}$ obtained by articulating only this joint to its lower and upper limits.
These quantities are visualized in \cref{fig:architecture}.
The decoder $\psi_\text{joint}$ is trained in a teacher-forcing style:
during training, query points are assigned to joints using their ground-truth part labels,
while during inference we use the segmentation prediction $\tilde{s}_j = \arg\max_p \tilde{\mathbf{S}}_{j,p}$.
Per-query results are aggregated by a geometric fitting step to recover one shared motion axis $\tilde{\mathbf{a}}_e$ and range $[\tilde{\theta}_e^-,\tilde{\theta}_e^+]$ for each joint in the kinematic tree $\mathcal{K}$.
We refer the reader to \cref{app:implementation-details} for more details.

\subsection{Training}

The model is trained end-to-end using a multi-task loss:
$
\mathcal{L} = \mathcal{L}_\mathrm{part} + \mathcal{L}_\mathrm{joint}
$,
where
$
\mathcal{L}_\mathrm{part} = \frac{1}{M}\sum_{j=1}^M \operatorname{CE}(\tilde{S}_j, m(\mathbf{q}_j))
$
is the cross-entropy loss for the part segmentation, and
$
\mathcal{L}_\mathrm{joint} = \frac{1}{P-1}\sum_{e=1}^{P-1}\sum_{\tau\in\tau_e}\sum_{m(\mathbf{q}_j)=v_e}\bigl(\lambda_d\|\tilde{\mathbf{d}}_{j,e}^{\tau}-\mathbf{d}_e^{\tau}\|_1+\lambda_c\|\tilde{\mathbf{c}}_{j,e}^{\tau}-\mathbf{c}_{j,e}^{\tau}\|_1+\lambda_q\|\tilde{\mathbf{q}}_{j,e}^{\tau,-}-\mathbf{q}_{j,e}^{\tau,-}\|_1+\lambda_q\|\tilde{\mathbf{q}}_{j,e}^{\tau,+}-\mathbf{q}_{j,e}^{\tau,+}\|_1\bigr)
$
supervises the over-parameterized joint targets.

\section{Experiments}%
\label{sec:exp}

We organize our experiments around three questions:
(1) \Cref{sec:exp-comparisons}: How does \method compare with existing methods for 3D articulation estimation and generation?
(2) \Cref{sec:exp-ablations}: How much does each component of the curated data mixture, and each conditioning modality, contribute to performance?
(3) \Cref{sec:exp-kinematic-prompting}: What capabilities does our model enable?

\subsection{Comparisons}%
\label{sec:exp-comparisons}

\begin{table}[t]
\centering
\small
\caption{
\textbf{Quantitative comparison}.
We compare against $8$ existing methods on the Lightwheel dataset~\cite{simready2025}.
For fairness, we evaluate \method under the same input mode as each baseline.
\method substantially outperforms all baselines.
\faCamera: image; \faCube: mesh; \faSitemap: kinematic condition; \faMapMarker*: point prompts.
Colors: \protect\best{best} and \protect\secondbest{second best}.
}%
\label{tab:comparison-notextfix}%
\label{tab:quantitative-comparison}
\setlength{\tabcolsep}{4pt}
\renewcommand{\arraystretch}{1.12}
\resizebox{\linewidth}{!}{%
\begin{tabular}{@{}lcccccccccccccc@{}}
\toprule
 & \multicolumn{4}{c}{\textbf{Input}} & \multicolumn{2}{c}{\textbf{Part Match}} & \multicolumn{3}{c}{\textbf{Rest-Pose Seg.}} & \multicolumn{3}{c}{\textbf{Articulated Geom.}} & \multicolumn{2}{c}{\textbf{Joint Axes}} \\
\cmidrule(lr){2-5}\cmidrule(lr){6-7}\cmidrule(lr){8-10}\cmidrule(lr){11-13}\cmidrule(lr){14-15}
\textbf{Method} & \faCamera & \faCube & \faSitemap & \faMapMarker* & Prec.$\uparrow$ & Rec.$\uparrow$ & gIoU$\uparrow$ & PC$\downarrow$ & mIoU$\uparrow$ & gIoU$\uparrow$ & PC$\downarrow$ & OC$\downarrow$ & AE$\downarrow$ & LE$\downarrow$ \\
\midrule
SINGAPO~\cite{liusingapo} & \checkmark &  &  &  & $37.6\%$ & \secondbest{$25.4\%$} & $-0.259$ & \secondbest{$0.200$} & $0.182$ & $-0.263$ & \secondbest{$0.295$} & \secondbest{$0.030$} & \secondbest{$30.3$} & \best{$0.077$} \\
PAct~\cite{liu2026pact} & \checkmark &  &  &  & $24.6\%$ & $16.8\%$ & $-0.430$ & $0.230$ & $0.092$ & $-0.434$ & $0.404$ & $0.056$ & $42.9$ & $0.188$ \\
PhysX-Anything~\cite{physxanything} & \checkmark &  &  &  & $20.3\%$ & $19.2\%$ & $-0.458$ & $0.247$ & $0.093$ & $-0.459$ & $0.334$ & $0.064$ & $38.8$ & \secondbest{$0.123$}  \\
URDF-Anything+~\cite{wu2026urdfanythingplus} & \checkmark &  &  &  & \secondbest{$70.7\%$} & $23.9\%$ & \secondbest{$-0.133$} & $0.259$ & \secondbest{$0.260$} & \secondbest{$-0.138$} & $0.305$ & $0.033$ & $50.4$ & $0.128$ \\
\textbf{\method{}} & \checkmark &  &  &  & \best{$73.4\%$} & \best{$57.6\%$} & \best{$0.177$} & \best{$0.102$} & \best{$0.405$} & \best{$0.164$} & \best{$0.161$} & \best{$0.015$} & \best{$18.1$} & $0.139$ \\
\midrule
Articulate AnyMesh~\cite{qiu2025articulate} &  & \checkmark &  &  & $89.2\%$ & $42.5\%$ & $0.172$ & $0.190$ & $0.452$ & $0.158$ & $0.237$ & $0.010$ & $21.7$ & $0.043$ \\
PartField~\cite{liu2025partfield} &  & \checkmark & \# &  & $61.2\%$ & \secondbest{$62.6\%$} & $0.079$ & \secondbest{$0.106$} & $0.264$ & --- & --- & --- & --- & --- \\
URDF-Anything+~\cite{wu2026urdfanythingplus} & (\checkmark) & \checkmark &  &  & $77.5\%$ & $27.4\%$ & $-0.104$ & $0.241$ & $0.267$ & $-0.110$ & $0.298$ & $0.026$ & $49.9$ & $0.111$ \\
Particulate~\cite{li2026particulate} &  & \checkmark &  &  & \secondbest{$89.9\%$} & $51.5\%$ & \secondbest{$0.332$} & $0.168$ & \secondbest{$0.576$} & \secondbest{$0.305$} & \secondbest{$0.208$} & \secondbest{$0.009$} & \secondbest{$20.9$} & \secondbest{$0.040$} \\
\textbf{\method{}} &  & \checkmark &  &  & \best{$94.3\%$} & \best{$74.8\%$} & \best{$0.583$} & \best{$0.091$} & \best{$0.724$} & \best{$0.542$} & \best{$0.108$} & \best{$0.004$} & \best{$13.9$} & \best{$0.018$} \\
\midrule
P3SAM~\cite{ma2025p3sam} &  & \checkmark & \# & \checkmark & \secondbest{$41.6\%$} & \secondbest{$79.7\%$} & \secondbest{$0.122$} & \secondbest{$0.177$} & \secondbest{$0.411$} & --- & --- & --- & --- & --- \\
\textbf{\method{}} &  & \checkmark & \checkmark & \checkmark & \best{$97.3\%$} & \best{$95.9\%$} & \best{$0.799$} & \best{$0.010$} & \best{$0.821$} & \best{$0.747$} & \best{$0.039$} & \best{$0.006$} & \best{$9.5$} & \best{$0.015$} \\
\bottomrule
\end{tabular}%
}
\end{table}

\paragraph{Baselines.}

We compare our model against eight existing methods with different input modalities:
SINGAPO~\cite{liusingapo}, PAct~\cite{liu2026pact}, PhysX-Anything~\cite{physxanything}, and URDF-Anything+~\cite{wu2026urdfanythingplus} take a single image as input and generate an articulated 3D object;
Articulate AnyMesh~\cite{qiu2025articulate} and Particulate~\cite{li2026particulate} take a 3D mesh as input and predict its articulated structure;
in addition, PartField~\cite{liu2025partfield} and P3SAM~\cite{ma2025p3sam} are two feed-forward 3D part segmentation methods.
Since URDF-Anything+ can optionally condition on a 3D mesh, we additionally compare against it using the ground-truth 3D object as input.
PartField and P3SAM are given the exact number of ground-truth parts at test time.

\paragraph{Evaluation protocol.}

We evaluate \method on the challenging Lightwheel dataset~\cite{simready2025} introduced by~\cite{li2026particulate}, which contains $243$ high-quality articulated objects spanning $14$ categories, including categories absent from prior datasets, such as stand mixers, range hoods, and cooktop stoves.
To align the conditioning setup with the baselines, we evaluate \method under three input modes:
(1) \emph{Image Only}, which provides only a rendered image of the object;
(2) \emph{Mesh}, which provides only the ground-truth 3D mesh $M$; and
(3) \emph{Mesh + Kinematic}, which additionally provides the ground-truth kinematic structure $\mathcal{K}$, along with per-part text $t_p$ and point prompts $\mathbf{x}_p$.
In the \emph{Image Only} setting, we first reconstruct a textured 3D mesh from the input image using the off-the-shelf 3D generator HY3D-3.1~\cite{tencent2026hunyuan3d31}.
For the \emph{Image Only} and \emph{Mesh} settings, we infer the kinematic condition $C$ with a VLM~\cite{googledeepmind2026gemini}, following a pipeline similar to \cref{sec:pseudo-labeling}.
We also prompt the VLM to localize a 2D point on each identified kinematic part, and unproject these points onto the corresponding input mesh (generated for \emph{Image Only}, ground-truth for \emph{Mesh}) to construct the point prompts.
Predicted and ground-truth parts have unknown correspondence and may differ in number (\ie, all baselines and our method in the \emph{Image Only} and \emph{Mesh} settings).
We thus perform Hungarian matching between the two sets based on pairwise part-centroid distances.

\paragraph{Metrics.}

We organize the evaluation metrics into four groups.
\emph{Part Match} assesses whether the predicted parts match the ground-truth parts in both granularity and spatial layout, using \emph{precision} and \emph{recall}.
We consider a matched pair of predicted and ground-truth parts to be valid if its generalized Intersection over Union~\cite{rezatofighi2019generalized} (gIoU) is at least $0$.
\emph{Rest-Pose Segmentation} measures segmentation quality in the rest pose using part-wise \emph{gIoU}, mean Intersection over Union (\emph{mIoU}), and bidirectional Chamfer distance (\emph{PC}), following~\cite{li2026particulate}.
\emph{Articulated Geometry} evaluates the geometry after fully articulating the predicted asset by moving every movable joint to its (predicted) upper-limit pose.
Following~\cite{li2026particulate}, we report part-wise \emph{gIoU}, part-wise Chamfer distance (\emph{PC}), and whole-object Chamfer distance (\emph{OC}), where these metrics provide a heuristic assessment of both part segmentation and joint motion estimation.
Finally, \emph{Joint Axes} directly measures joint-axis accuracy using angle error (\emph{AE}) and location error (\emph{LE}) between matched predicted and ground-truth joints.

\paragraph{Results.}

\begin{figure}[t]
\centering
\includegraphics[width=\linewidth]{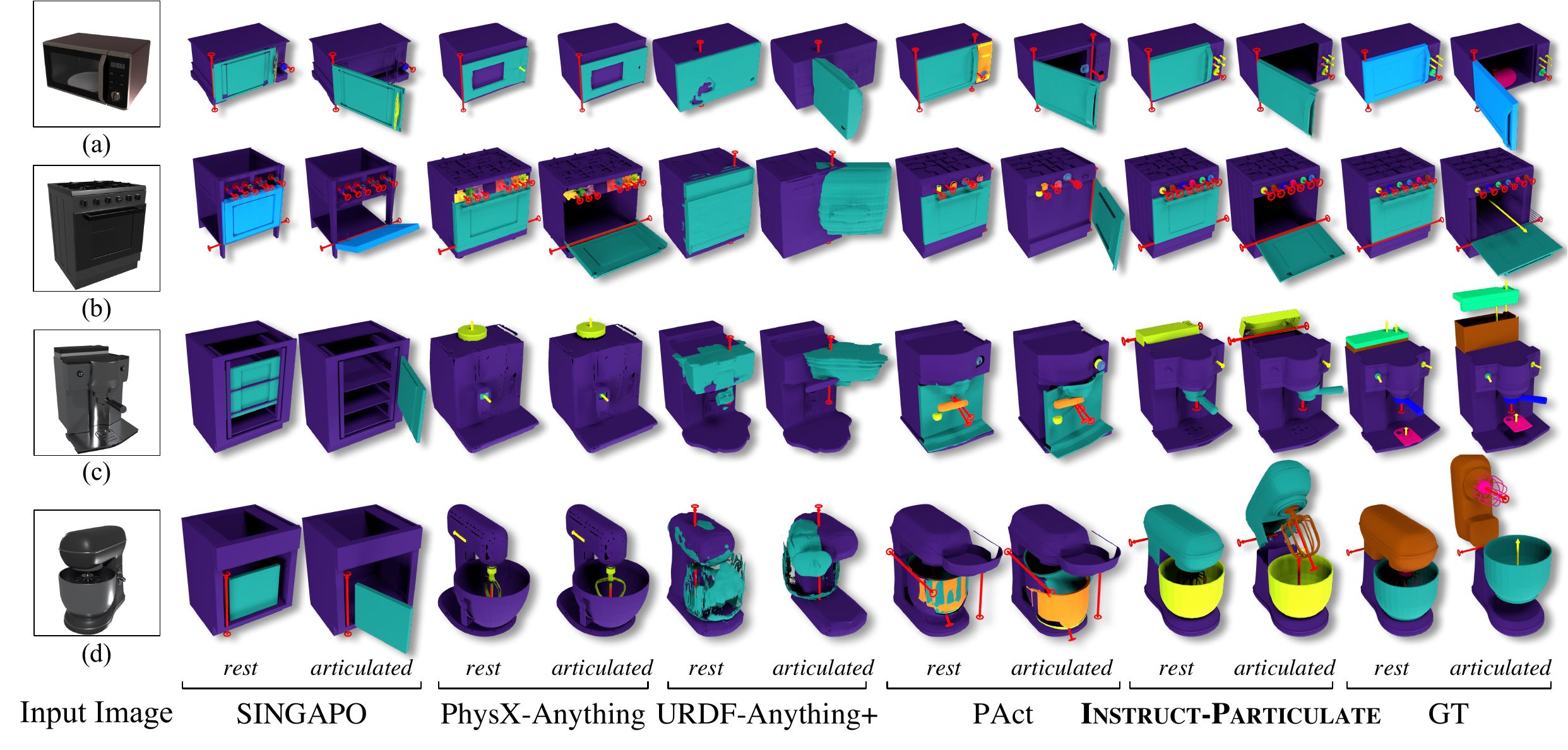}
\vspace{-2em}
\caption{
\textbf{Qualitative comparison} (\emph{Image Only} mode).
Given a single image as input, \method (combined with an off-the-shelf 3D generator) can generate more realistic articulated 3D objects than the baselines.
}%
\label{fig:qualitative-comparison}
\end{figure}

We report quantitative results in~\cref{tab:quantitative-comparison}, where \method consistently outperforms all baselines across all input settings.
\Cref{fig:qualitative-comparison} shows image-conditioned articulated 3D generation results from our approach and the baselines on the Lightwheel benchmark.
Even for common categories, such as the microwave oven in (a) and the stove in (b), most baselines fail to reliably recover small articulated parts, including buttons and knobs.
On more complex objects, such as the coffee machine in (c) and the stand mixer in (d), methods such as SINGAPO and URDF-Anything+ fail entirely.
By contrast, \method is robust to synthetic meshes, allowing us to offload geometry reconstruction to an off-the-shelf 3D generator, while recovering small articulated parts and estimating their joint motion accurately.
We provide qualitative comparisons in the \emph{Mesh} setting in \cref{app:additional-results}.

\subsection{Ablation Studies}%
\label{sec:exp-ablations}

\begin{table}[t]
\centering
\small
\caption{
\textbf{Data ablations}.
We incrementally add curated data sources from~\cref{sec:scaling-data} (%
PM\@: PartNet-Mobility,
GRS\@: GRScenes).
The added sources yield complementary gains in part segmentation and joint motion estimation.
Colors: \protect\best{best} and \protect\secondbest{second best}.
}%
\label{tab:data-ablations}
\setlength{\tabcolsep}{2pt}
\renewcommand{\arraystretch}{1.12}
\begin{tabular}{@{}ccccccccccccccc@{}}
\toprule
& \multicolumn{4}{c}{\textbf{Training Data}} & \multicolumn{2}{c}{\textbf{Part Match}} & \multicolumn{3}{c}{\textbf{Rest-Pose Seg.}} & \multicolumn{3}{c}{\textbf{Articulated Geom.}} & \multicolumn{2}{c}{\textbf{Joint Axes}} \\
\cmidrule(lr){2-5}\cmidrule(lr){6-7}\cmidrule(lr){8-10}\cmidrule(lr){11-13}\cmidrule(lr){14-15}
& PM+GRS
& S.~\ref{sec:generic-part-segmented}
& S.~\ref{sec:pseudo-labeling}
& S.~\ref{sec:articulated-3d-models-with-coding-agents}
& Prec.$\uparrow$ & Rec.$\uparrow$ & gIoU$\uparrow$ & PC$\downarrow$ & mIoU$\uparrow$
& gIoU$\uparrow$ & PC$\downarrow$ & OC$\downarrow$ & AE$\downarrow$ & LE$\downarrow$ \\
\midrule
$\mathbb{A}$ & \checkmark & & & & $89.3\%$ & $71.2\%$ & $0.477$ & $0.085$ & $0.620$ & $0.446$ & $0.124$ & $0.009$ & $13.0$ & $0.036$ \\
$\mathbb{B}$ & \checkmark & \checkmark & & & $95.3\%$ & $89.2\%$ & $0.692$ & $0.032$ & $0.743$ & $0.652$ & $0.070$ & \secondbest{$0.008$} & \secondbest{$12.3$} & $0.025$ \\
$\mathbb{C}$ & \checkmark & \checkmark & \checkmark & & \secondbest{$96.8\%$} & \secondbest{$95.3\%$} & \secondbest{$0.791$} & \best{$0.009$} & \secondbest{$0.813$} & \secondbest{$0.745$} & \secondbest{$0.049$} & \secondbest{$0.008$} & $13.1$ & \secondbest{$0.020$} \\
$\mathbb{D}$ & \checkmark & \checkmark & \checkmark & \checkmark & \best{$96.9\%$} & \best{$95.4\%$} & \best{$0.803$} & \secondbest{$0.010$} & \best{$0.825$} & \best{$0.757$} & \best{$0.041$} & \best{$0.007$} & \best{$11.0$} & \best{$0.018$} \\
\bottomrule
\end{tabular}%
\end{table}

\begin{table}[t]
\centering
\small
\caption{
\textbf{Conditioning modality ablations}.
We compare a model trained without kinematic conditioning ($\mathbb{A}$) with variants that disable point prompts $x_p$ ($\mathbb{B}$) or text prompts $t_p$ ($\mathbb{C}$) at inference time.
Both prompt types improve performance, while kinematic conditioning disambiguates the target structure and enables learning from diverse annotations.
\faLanguage: text prompts, \faMapMarker*: point prompts.
Colors: \protect\best{best} and \protect\secondbest{second best}.
}%
\label{tab:conditioning-ablations}
\setlength{\tabcolsep}{4pt}
\renewcommand{\arraystretch}{1.12}
\begin{tabular}{@{}cccccccccccccc@{}}
\toprule
& \multicolumn{1}{c}{\textbf{Train}} & \multicolumn{2}{c}{\textbf{Inference}} & \multicolumn{2}{c}{\textbf{Part Match}} & \multicolumn{3}{c}{\textbf{Rest-Pose Seg.}} & \multicolumn{3}{c}{\textbf{Articulated Geom.}} & \multicolumn{2}{c}{\textbf{Joint Axes}} \\
\cmidrule(lr){2-2}\cmidrule(lr){3-4}\cmidrule(lr){5-6}\cmidrule(lr){7-9}\cmidrule(lr){10-12}\cmidrule(lr){13-14}
& $C$
& \faLanguage
& \faMapMarker*
& Prec.$\uparrow$ & Rec.$\uparrow$ & gIoU$\uparrow$ & PC$\downarrow$ & mIoU$\uparrow$
& gIoU$\uparrow$ & PC$\downarrow$ & OC$\downarrow$ & AE$\downarrow$ & LE$\downarrow$ \\

\midrule
$\mathbb{A}$ & $\times$ & --- & --- & $64.1\%$ & $74.7\%$ & $0.249$ & $0.123$ & $0.463$ & $0.226$ & $0.163$ & $0.011$ & $22.9$ & $0.048$ \\
$\mathbb{B}$ & \checkmark & \checkmark & & $87.7\%$ & $66.5\%$ & $0.425$ & $0.096$ & $0.576$ & $0.398$ & $0.132$ & \best{$0.007$} & \secondbest{$13.8$} & $0.079$ \\
$\mathbb{C}$ & \checkmark & & \checkmark & \secondbest{$95.6\%$} & \secondbest{$91.1\%$} & \secondbest{$0.626$} & \secondbest{$0.028$} & \secondbest{$0.663$} & \secondbest{$0.594$} & \secondbest{$0.064$} & \best{$0.007$} & $18.8$ & \secondbest{$0.035$} \\
$\mathbb{D}$ & \checkmark & \checkmark & \checkmark & \best{$96.9\%$} & \best{$95.4\%$} & \best{$0.803$} & \best{$0.010$} & \best{$0.825$} & \best{$0.757$} & \best{$0.041$} & \best{$0.007$} & \best{$11.0$} & \best{$0.018$} \\
\bottomrule
\end{tabular}%
\end{table}

\paragraph{Data.}

We train separate models using identical hyperparameters on different data mixtures, starting from the existing articulated 3D datasets PartNet-Mobility~\cite{xiang20sapien:} and GRScenes~\cite{wang2024grutopia} used in~\cite{li2026particulate}, and incrementally adding the curated data sources described in~\cref{sec:pseudo-labeling,sec:generic-part-segmented,sec:articulated-3d-models-with-coding-agents}.
In \cref{tab:data-ablations}, we report the results obtained by the best checkpoint from each training run.
Data generated by the coding agent with full joint parameter supervision (\cref{sec:articulated-3d-models-with-coding-agents}) mainly contributes to the \emph{Joint Axes} metrics ($\mathbb{C}$ \vs $\mathbb{D}$),
while the larger-scale part-segmented data from~\cref{sec:generic-part-segmented,sec:pseudo-labeling} improves part segmentation quality ($\mathbb{A}$ \vs $\mathbb{B}$ \vs $\mathbb{C}$).
The model trained with only existing datasets (\ie, $\mathbb{A}$) severely overfits to the available shapes and kinematic structures, and struggles to generalize to new categories.

\paragraph{Conditioning.}

We further ablate the conditioning modalities in~\cref{tab:conditioning-ablations}.
Using the same checkpoint trained on the full data mixture, we disable point prompts $\mathbf{x}_p$ ($\mathbb{B}$) or text prompts $t_p$ ($\mathbb{C}$) at inference time.
Both modalities provide useful context for articulation estimation, with point prompts providing the larger gain.
We also train a separate model without kinematic conditioning ($\mathbb{A}$), using the architecture of~\cite{li2026particulate} and supervising parts after Hungarian matching between predicted and ground-truth parts.
Its much weaker performance suggests that, without explicit conditioning, variation in part granularity and semantics across datasets encourages the model to average over multiple plausible annotations and produce suboptimal results.

\subsection{Articulated 3D Object Generation and Kinematic Prompting}%
\label{sec:exp-kinematic-prompting}
\begin{figure}[t]
\centering
\includegraphics[width=\linewidth]{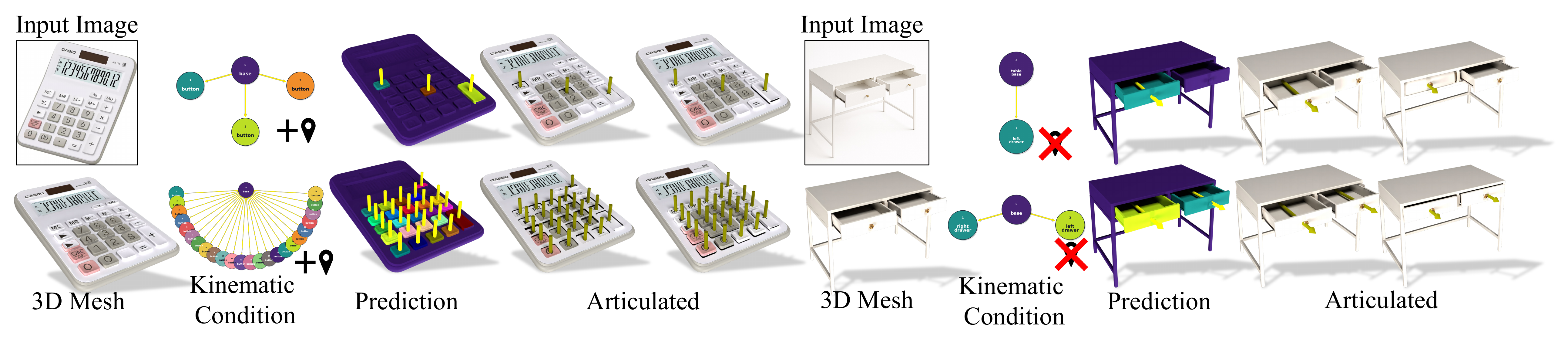}
\vspace{-2em}
\caption{
\textbf{Qualitative results with kinematic prompting}.
We show articulated structures predicted by our model given different kinematic conditions.
The model follows the specification faithfully.
}%
\label{fig:qualitative-results}
\end{figure}

\paragraph{Image-conditioned articulated 3D object generation.}

While our model takes existing 3D assets as input, it is robust to AI-generated meshes, and thus enables image-conditioned articulated 3D object generation via an off-the-shelf 3D generator.
All results shown in \cref{fig:teaser} are obtained in this manner.
This pipeline can support the creation of diverse simulation assets for embodied AI training.

\paragraph{Kinematic prompting.}

Beyond disambiguating outputs, kinematic conditioning also enables test-time prompting.
In \cref{fig:qualitative-results}, we show predictions for the same input mesh under different kinematic conditions.
The model follows these specifications faithfully, including a challenging case with $24$ button joints.
It also supports text-guided spatial control: for the desk example, the model receives no point prompts to distinguish the two drawers, but instead uses spatial text prompts (\ie, ``left drawer'' and ``right drawer'').
This reflects our data-curation effort to disambiguate spatial descriptors (\cref{sec:generic-part-segmented}), allowing reliable spatial reasoning when the object orientation is well defined.

\section{Conclusions}

We have presented \method, a feed-forward model for recovering the articulated structure of a static 3D object, and endowed it with kinematic control.
Our model is built on an efficient encoder-decoder architecture and is trained on more than $150$k 3D objects, for which we developed a new data annotation engine that can label 3D articulated parts using a VLM\@.
In this way, \method generalizes much better to novel and more diverse objects than prior works.
Furthermore, when paired with an off-the-shelf 3D generator, \method provides a practical pipeline to generate articulated 3D assets directly from real-world images.

\section*{Acknowledgements}
Ruining Li is supported by a Toshiba Research Studentship.
Chuanxia Zheng is supported by NTU SUG-NAP and National Research Foundation, Singapore, under its NRF Fellowship Award NRF-NRFF17-2025-0009.
Christian Rupprecht is supported by an Amazon Research Award and ERC StG Volute (grant no. 101222037).
This work is partially supported by the UKRI AIRR programme (ID: u6en)
and ERC CoG 101001212-UNION\@.

{
    \small
    \bibliographystyle{plainnat}
    \bibliography{ref,vedaldi_general,vedaldi_specific}
}

\newpage
\appendix
\section{Implementation Details}%
\label{app:implementation-details}

\subsection{Training Details}

\paragraph{Data mixture.}

Our final \method model is trained on the available articulated 3D datasets PartNet-Mobility~\cite{xiang20sapien:} and GRScenes~\cite{wang2024grutopia}, together with the curated data sources introduced in~\cref{sec:pseudo-labeling,sec:generic-part-segmented,sec:articulated-3d-models-with-coding-agents}.
The training data mixture is determined empirically and is summarized in~\cref{tab:sup-data-mixture}.

\begin{table}[h]
\centering
\small
\caption{
\textbf{Training data mixture} of \method.
}%
\label{tab:sup-data-mixture}
\setlength{\tabcolsep}{5pt}
\renewcommand{\arraystretch}{1.12}
\begin{tabular}{@{}llcc@{}}
\toprule
\textbf{Source} & \textbf{Type} & \textbf{Scale} & \textbf{Weight} \\
\midrule
PartNet-Mobility~\cite{xiang20sapien:} & Existing articulated assets & $2.3$k & $0.17$ \\
GRScenes~\cite{wang2024grutopia} & Existing articulated assets & $1.8$k & $0.10$ \\
HY3D-Bench-Part-Level~\cite{hunyuan3d2026hy3dbench} & Generic part-segmented assets & $117$k & $0.18$ \\
HY3D-Bench-Synthetic~\cite{hunyuan3d2026hy3dbench} & Synthetic assets pseudo-labeled with kinematic parts & $27$k & $0.22$ \\
Articraft~\cite{zhou2026articraft} & Agent-generated articulated assets & $10$k & $0.33$ \\
\bottomrule
\end{tabular}%
\end{table}

\paragraph{Data augmentation.}
During training, we sample a random articulated state for each training iteration.
All input shapes are first normalized to $[-0.5, 0.5]^3$, and then rotated around the $z$-axis by a random angle sampled from $\{0, \pi / 2, \pi, 3\pi / 2\}$.
We further apply a random scale factor drawn from $\mathcal{U}(0.95, 1.05)$ and a random translation vector from $\mathcal{N}(0, 0.05)^3$.
We randomly remove normals with a probability of $0.3$.
For each training iteration, we also randomly merge each part to its parent part in the kinematic tree with a probability of $0.15$.
The text prompt $t_p$ is randomly removed with a probability of $0.2$ independently for each part in the (augmented) kinematic tree $\mathcal{K}$,
while the point prompt $x_p$ is randomly removed with a probability of $0.25$.
We make sure the point prompt and text prompt are not both removed for the same part.

\paragraph{Hyperparameters.}
The main training hyperparameters are summarized in~\cref{tab:sup-hyperparameters}.

\begin{table}[h]
\centering
\small
\caption{
\textbf{Training configuration} of \method.
}%
\label{tab:sup-hyperparameters}
\setlength{\tabcolsep}{4pt}
\renewcommand{\arraystretch}{1.12}
\begin{tabular}{@{}l c@{}}
\toprule
\textbf{Hyperparameter name} & \textbf{Value} \\
\midrule
\multicolumn{2}{@{}l}{\textbf{Model}} \\
Number of attention blocks $B$ & $6$ \\
Number of shape points $N$ & $10{,}240$ \\
Number of latent shape tokens $L$ & $2{,}048$ \\
Model dimension $D$ & $768$ \\
\midrule
\multicolumn{2}{@{}l}{\textbf{Loss weights}} \\
$\lambda_d$ & $5$ \\
$\lambda_c$ & $5$ \\
$\lambda_q$ & $10$ \\
\midrule
\multicolumn{2}{@{}l}{\textbf{Optimization}} \\
Optimizer & AdamW \\
Peak learning rate & $6.4{\times}10^{-5}$ \\
Warmup steps & $1{,}000$ \\
Global batch size & $128$ \\
Number of query points decoded together $M$ & $4{,}096$ \\
Training steps & $100{,}000$ \\
Training compute & $16$ NVIDIA Blackwell GPUs \\
Weight decay & $0.01$ \\
\bottomrule
\end{tabular}%
\end{table}

\subsection{Inference Details}

\paragraph{Part segmentation inference.}

We perform inference with a much denser query point cloud ($102,400$ points \vs $4,096$ points during training) to ensure sufficient coverage of the object surface.
We obtain the part labels from the predicted logits as $\tilde{s}_j = \arg\max_p \tilde{\mathbf{S}}_{j,p}$.

\paragraph{Optimizing joint motion constraints from over-parameterized targets.}

Once we obtain the per-query-point over-parameterized joint motion targets $\tilde{\mathbf{d}}_{j,e}^{\tau}$, $\tilde{\mathbf{c}}_{j,e}^{\tau}$, $\tilde{\mathbf{q}}_{j,e}^{\tau,-}$, and $\tilde{\mathbf{q}}_{j,e}^{\tau,+}$ from the decoder,
we run an optimization to recover the joint axis $\mathbf{a}_e = (\mathbf{d}_e, \mathbf{p}_e)$ and range $[\theta_e^-,\theta_e^+]$ for each joint $e$.
Specifically, we first aggregate each query point's ``vote'' for the joint direction, $\tilde{\mathbf{d}}_{j,e}^{\tau}$, and pivot point, $\tilde{\mathbf{c}}_{j,e}^{\tau}$,
to obtain the joint axis
$
(
    \frac{1}{|\mathcal{Q}(e)|}\sum_{j\in\mathcal{Q}(e)} \tilde{\mathbf{d}}_{j,e}^{\tau},
    \frac{1}{|\mathcal{Q}(e)|}\sum_{j\in\mathcal{Q}(e)} \tilde{\mathbf{c}}_{j,e}^{\tau}
)
$,
where $\mathcal{Q}(e) = \{j \mid \tilde{s}_j = v_e\}$ is the set of query points predicted to belong to the child part $v_e$.
We then solve the motion bounds $[\theta_e^-,\theta_e^+]$ as
$
\arg\min_{\theta_e^-,\theta_e^+} \sum_{j\in\mathcal{Q}(e)} \left \| \tilde{\mathbf{q}}_{j,e}^{\tau,-} - F(\mathbf{q}_j, \mathbf{a}_e, \theta_e^{-}, \tau_e) \right \|_2 + 
\left \| \tilde{\mathbf{q}}_{j,e}^{\tau,+} - F(\mathbf{q}_j, \mathbf{a}_e, \theta_e^+, \tau_e) \right \|_2,
$
where $F(\mathbf{q}_j, \mathbf{a}_e, \theta_e, \tau_e)$ is the forward kinematics function that computes the location of a query point $\mathbf{q}_j$ given the joint axis $\mathbf{a}_e$ and motion bounds $[\theta_e^-,\theta_e^+]$.
Empirically, we find that first fitting the axis based on $\tilde{\mathbf{d}}_{j,e}^{\tau}$ and $\tilde{\mathbf{c}}_{j,e}^{\tau}$ yields better results than performing global optimization over both $\mathbf{a}_e$ and $[\theta_e^-,\theta_e^+]$.

\section{Additional Results}%
\label{app:additional-results}

\subsection{Additional Qualitative Comparisons}%
\label{app:additional-qualitative-comparisons}

\begin{figure}[t]
\centering
\includegraphics[width=\linewidth]{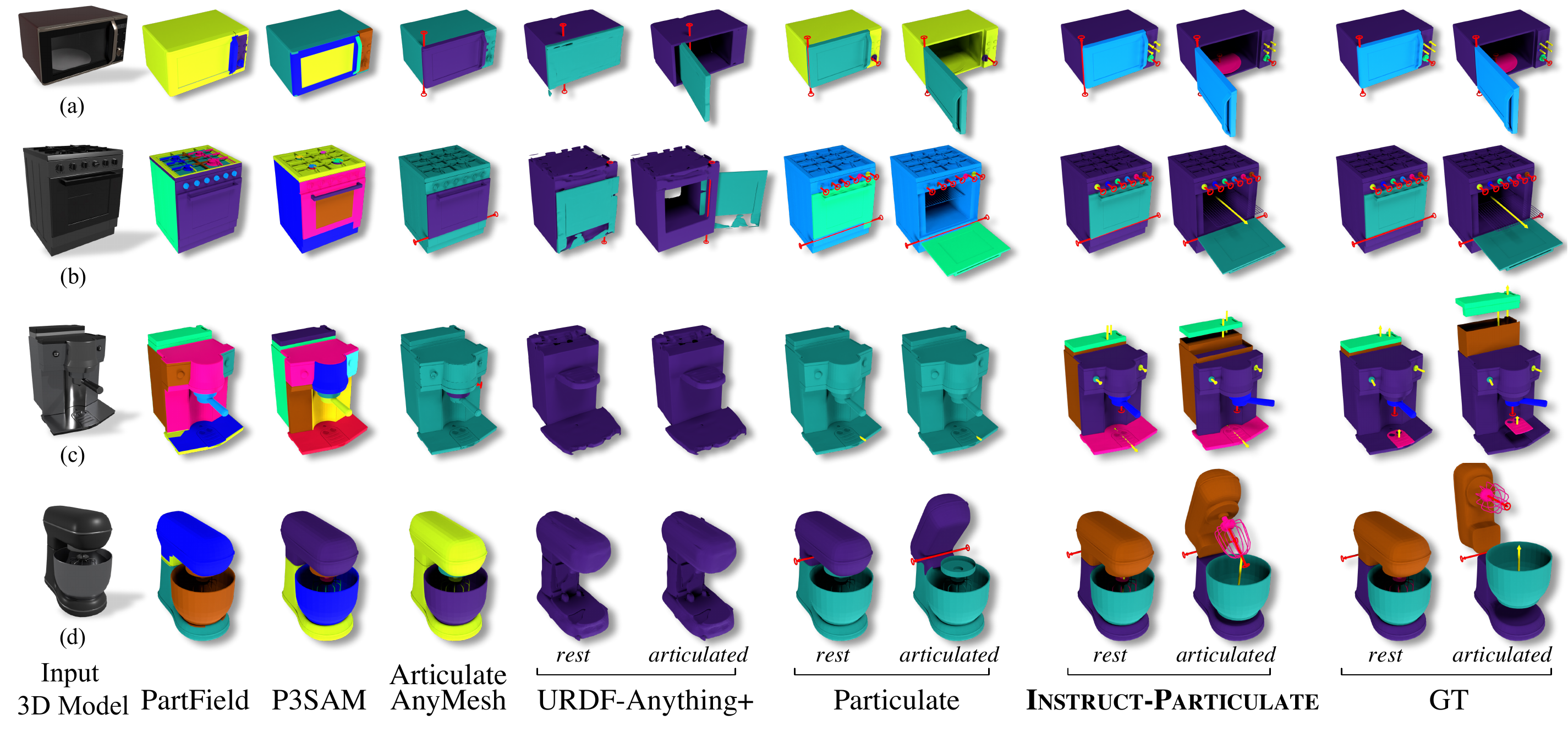}
\vspace{-2em}
\caption{
\textbf{Qualitative comparison} (\emph{Mesh} mode).
Given a 3D mesh as input, \method (combined with a VLM that labels the kinematic structure) can more reliably segment small and internal parts than the baselines, while generalizing to more complex objects.
}%
\label{fig:qualitative-comparison-mesh}
\end{figure}

In~\cref{fig:qualitative-comparison-mesh}, we present additional qualitative comparisons in the \emph{Mesh} setting, where each method takes an artist-created 3D mesh from the Lightwheel benchmark as input.
PartField and P3SAM are designed for \emph{semantic} part segmentation, whose part definitions often do not align with the \emph{articulated} parts required for kinematic reasoning.
While baseline methods can produce plausible results on common categories, such as the microwave oven in (a) and the stove in (b), they generalize less reliably to more complex objects, such as the coffee machine in (c) and the stand mixer in (d),
and often miss small and internal parts.
By contrast, \method reliably segments them, including the microwave oven's rotating plate and the buttons and knobs of the stove and coffee machine, while also estimating their joint motion accurately.

\subsection{Failure Cases and Limitations}%
\label{app:limitations-and-failure-cases}
\begin{figure}[t]
\centering
\includegraphics[width=\linewidth]{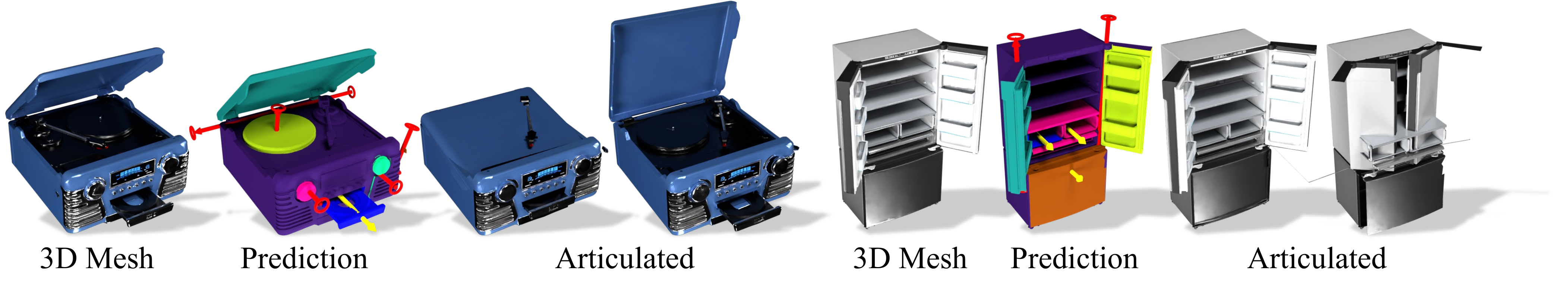}
\vspace{-2em}
\caption{
\textbf{Failure cases} of \method.
}%
\label{fig:failure-cases}
\end{figure}

\paragraph{Failure cases.}

We present representative failure cases in \cref{fig:failure-cases}.
While our model is generally robust to AI-generated meshes, segmentation artifacts can still occur, such as on the right knob of the CD player.
Because each joint axis is estimated by aggregating votes from query points on the predicted part, such local errors can propagate to joint-motion estimation.
Artifacts in the generated meshes can also degrade prediction quality, as shown by the floating component in the CD player and the missing part separation in the refrigerator.

\paragraph{Limitations.}

While \method can support large-scale creation of simulation assets, its outputs are \emph{not} yet simulation-ready.
They lack physical properties, and AI-generated meshes may contain incomplete geometry or excessive face counts.
Improving the simulation readiness of these assets (\eg, via post-training~\cite{li25dso}), remains future work.

\end{document}